\relax
\documentclass[letterpaper]{article} 
\usepackage{aaai22}  
\usepackage{times}  
\usepackage{helvet}  
\usepackage{courier}  
\usepackage[hyphens]{url}  
\usepackage{graphicx} 
\usepackage{natbib}  
\usepackage{caption} 
\DeclareCaptionStyle{ruled}{labelfont=normalfont,labelsep=colon,strut=off} 
\usepackage{algorithm}
\usepackage{algorithmic}

\usepackage{newfloat}
\usepackage{listings}
\floatstyle{ruled}
\newfloat{listing}{tb}{lst}{}
\floatname{listing}{Listing}
\usepackage{adjustbox}
\usepackage{booktabs}
\usepackage{amsmath,amssymb,amsthm}
\usepackage{caption,subcaption,float}

\title{Industry-Scale Orchestrated Federated Learning for Drug Discovery}
\author{
   Martijn Oldenhof\textsuperscript{\rm 1}, Gergely \'Acs\textsuperscript{\rm 2}, Bal\'azs Pej\'o\textsuperscript{\rm 2}, Ansgar Schuffenhauer\textsuperscript{\rm 4}, Nicholas Holway\textsuperscript{\rm 4}, Noé Sturm\textsuperscript{\rm 4}, Arne Dieckmann\textsuperscript{\rm 5}, Oliver Fortmeier\textsuperscript{\rm 5}, Eric Boniface\textsuperscript{\rm 6}, Clément Mayer\textsuperscript{\rm 6}, Arnaud Gohier\textsuperscript{\rm 8}, Peter Schmidtke\textsuperscript{\rm 7}, Ritsuya Niwayama\textsuperscript{\rm 8}, Dieter Kopecky\textsuperscript{\rm 9}, Lewis Mervin\textsuperscript{\rm 10}, Prakash Chandra Rathi\textsuperscript{\rm 11}, Lukas Friedrich\textsuperscript{\rm 14}, András Formanek\textsuperscript{\rm 1, \rm 3}, Peter Antal\textsuperscript{\rm 3}, Jordon Rahaman\textsuperscript{\rm 16}\footnote{Work was carried out as employee of Amgen.}, Adam Zalewski\textsuperscript{\rm 15}, Wouter Heyndrickx\textsuperscript{\rm 17}, Ezron Oluoch\textsuperscript{\rm 18}, Manuel Stößel\textsuperscript{\rm 18}, Michal Vančo\textsuperscript{\rm 18}, David Endico\textsuperscript{\rm 19}, Fabien Gelus\textsuperscript{\rm 19}, Thaïs de Boisfossé\textsuperscript{\rm 19}, Adrien Darbier\textsuperscript{\rm 19}, Ashley Nicollet\textsuperscript{\rm 19}, Matthieu Blottière\textsuperscript{\rm 19}, Maria Telenczuk\textsuperscript{\rm 19}, Van Tien Nguyen\textsuperscript{\rm 19}, Thibaud Martinez\textsuperscript{\rm 19}, Camille Boillet\textsuperscript{\rm 19}, Kelvin Moutet\textsuperscript{\rm 19}, Alexandre Picosson\textsuperscript{\rm 19}, Aurélien Gasser\textsuperscript{\rm 19}, Inal Djafar\textsuperscript{\rm 19}, Antoine Simon\textsuperscript{\rm 19}, Ádám Arany\textsuperscript{\rm 1}, Jaak Simm\textsuperscript{\rm 1}, Yves Moreau\textsuperscript{\rm 1}, Ola Engkvist\textsuperscript{\rm 12,13}, Hugo Ceulemans\textsuperscript{\rm 17}, Camille Marini\textsuperscript{\rm 19} and Mathieu Galtier\textsuperscript{\rm 19}
}
\affiliations{

\textsuperscript{\rm 1}KU Leuven, ESAT-STADIUS, Kasteelpark Arenberg 10, 3001 Heverlee, Belgium. \\
\textsuperscript{\rm 2}BME-HIT, CrySyS Lab, Budapest, Hungary. \\
\textsuperscript{\rm 3}BME-MIT, Budapest, Hungary.\\
\textsuperscript{\rm 4}Novartis Institutes for BioMedical Research, Novartis Campus, CH-4002 Basel, Switzerland.\\
\textsuperscript{\rm 5}Bayer AG, 51373 Leverkusen, Germany.\\
\textsuperscript{\rm 6}Substra Foundation - Labelia Labs, Nantes, France.\\ \textsuperscript{\rm 7}Discngine, 79 Avenue Ledru Rollin, 75012 Paris, France.\\
\textsuperscript{\rm 8}Institut de recherches Servier, 125 chemin de ronde, 78290 Croissy-sur-Seine, France.\\
\textsuperscript{\rm 9}Boehringer Ingelheim RCV GmbH \& Co KG, Vienna, Austria.\\
\textsuperscript{\rm 10}Molecular AI, Discovery Sciences, R\&D, AstraZeneca, Cambridge, UK.\\ 
\textsuperscript{\rm 11}R\&D IT, AstraZeneca, Cambridge, UK.\\
\textsuperscript{\rm 12}Molecular AI, Discovery Sciences, R\&D, AstraZeneca, Gothenburg, Sweden.\\
\textsuperscript{\rm 13}Department of Computer Science and Engineering, Chalmers University of Technology, Gothenburg, Sweden.\\
\textsuperscript{\rm 14}Merck KGaA, Global Research \& Development, Frankfurter Strasse 250, 64293 Darmstadt, Germany.\\
\textsuperscript{\rm 15}Amgen Research (Munich) GmbH, Staffelseestraße 2, 81477, Munich, Germany.\\
\textsuperscript{\rm 16}Pillar Biosciences, Inc., 9 Strathmore Rd Natick, MA 01760, USA.\\
\textsuperscript{\rm 17}Janssen Pharmaceutica NV, Turnhoutseweg 30, 2340 Beerse, Belgium.\\
\textsuperscript{\rm 18}Kubermatic, Hamburg, Germany. \\
\textsuperscript{\rm 19}Owkin, Paris, France.\\
martijn.oldenhof@kuleuven.be, mathieu.galtier@owkin.com

}

\begin{document}

\maketitle

\begin{abstract}
To apply federated learning to drug discovery we developed a novel platform in the context of European Innovative Medicines Initiative (IMI) project MELLODDY (grant n°831472), which was comprised of 10 pharmaceutical companies, academic research labs, large industrial companies and startups. The MELLODDY platform was the first industry-scale platform to enable the creation of a global federated model for drug discovery without sharing the confidential data sets of the individual partners. The federated model was trained on the platform by aggregating the gradients of all contributing partners in a cryptographic, secure way following each training iteration. The platform was deployed on an Amazon Web Services (AWS) multi-account architecture running Kubernetes clusters in private subnets. Organisationally, the roles of the different partners were codified as different rights and permissions on the platform and administrated in a decentralized way. The MELLODDY platform generated new scientific discoveries which are described in a companion paper.
\end{abstract}

\section{Introduction}

Billions of Euros in research and development are needed to successfully bring a new drug to the market. Furthermore, drug discovery and development is a high risk process as there is a failure rate of around 90\% for drug candidates that reach the clinical studies phase. Therefore, making the early stages of drug discovery more efficient and accurate holds the potential to have a significant impact on the pharmaceutical industry. 
    
Tools and models based on machine learning and artificial intelligence are commonly applied in all stages of drug discovery and development to make the process  more efficient. A standard technique is to use quantitative structure-activity relationship (QSAR) machine learning models~\cite{ghasemi2018neural} to predict bioactivity or toxicity of small molecules and possible drug candidates. However, the lack of data often limits model performance improvement. A collaborative approach that brings together industry competitors to leverage vast datasets holds the potential to overcome this challenge and enable better model performance.

Several levels of collaboration in terms of privacy and computational overhead can be envisioned. For data privacy, the focus is on the proper handling of sensitive data, including confidential data such as intellectual property, and on protecting the confidentiality and immutability of the data. Ensuring data privacy could lead to computational overhead caused by extra process requirements (e.g. authentication and encryption). The most efficient in computation usually means the least effective in privacy (e.g. a centralised server building a model by pooling the data sets from all partners would enable the models to profit from a large pool of data, however it would enable all or some partners accessing data of others). At the other extreme end lies cryptographic techniques such as secure multi-party computation (SMPC)~\cite{cramer2015secure} and homomorphic encryption~\cite{gentry2009fully}, that could increase the levels of privacy guarantees but are less practical when applied on big data use cases due to computational overhead. Federated learning~\cite{mcmahan2017communication}, by design, provides a minimal required level of privacy as the data stays under control of each participant while computational overhead is still reasonable for big data use cases.

\begin{figure}[t]
        \centering
        \includegraphics[width=0.9\columnwidth]{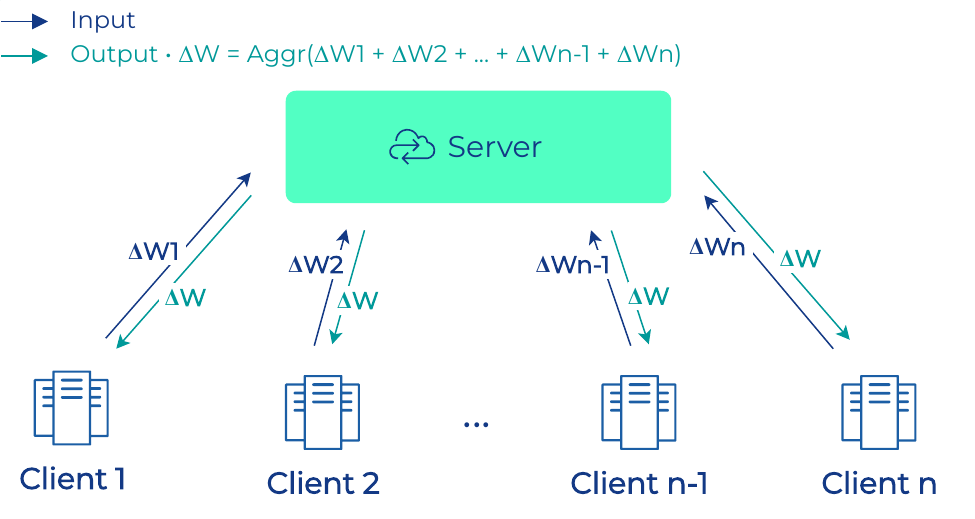}
        \caption{Illustrating Federated Learning: 
          within a single training round, the clients update the model (i.e. $W$) using their local training sets and share only the corresponding gradients (i.e. $\Delta W_i$) with the Server, who aggregates them (e.g. by averaging) and broadcast the resulting model update (i.e. $\Delta W$) back to the clients.}
        \label{fig:sec_aggr}
    \end{figure}
    
    Federated Learning has already been used in several application fields where the data are sensitive. It originally emerged~\cite{mcmahan2017communication} as part of edge computing with mobile applications for which user data were too sensitive to gather in a single place. It has also been used in the automotive industry and more extensively in healthcare where medical data are highly regulated and sometimes too valuable to share openly~\cite{LI2020106854}. There has been numerous academic publications about Federated Learning with virtually split data sets~\cite{mammen2021federated, li2020federated}, but the deployment of such technology in real life has been much scarcer. Industry-scale federated learning comes hand-in-hand with many challenges like scalability (computationally) as well as the synchronization of the data preparation and the orchestration of different partners operationally.

    In this paper, we describe a real world application for drug discovery in the context of a European Innovative Medicines Initiative (IMI) project called MELLODDY (grant n°831472), which gathered 10 pharmaceutical companies, academic research labs, large industrial companies and startups. The platform developed for MELLODDY project enabled to produce scientific results described in the work of~\citet{melloddy}. The data used in MELLODDY were chemical and assay data resulting from research conducted by ten pharmaceutical companies over the span of decades. The companies treat these data as trade secrets, as the vast majority of it is not yet protected by patents and therefore cannot be disclosed to the public or to competitors. Thus, no pharmaceutical company is willing to share its data with another. However, they are interested in sharing common predictive models which have been trained using a scheme visualised in Figure~\ref{fig:sec_aggr} on their combined data sets, provided their data sets can not be accessed or inferred by anyone.
\section{MELLODDY Use Case}
    
    \subsection{Data}\label{sec:data} 
    
    MELLODDY built machine-learning models on data resulting from in vitro bioassay measurements on small molecule samples in the early phases of drug discovery~\cite{Hughes2011}. The main prediction tasks were from assays run in concentration-response mode, where a compound is measured at multiple concentrations from which an AC50 value is derived~\cite{Beck2004}, typically by the use of automated curve fitting system~\cite{Gubler2018}. These data are very sparse; far less than 1\% of the structure activity matrix is filled. Single concentration high-throughput screening~\cite{Macarron2011} data is available in larger volumes, and can be used as auxiliary tasks to be included in the training, but ignored for performance evaluation purposes.
    In contrast to other efforts focusing on public data, where results of bioassays on the same target were merged into one task, for example ~\cite{Sturm2020} , each bioassay in MELLODDY was presented as its own task (or group of tasks). This means that there was little overlap between the prediction tasks of different pharmaceutical companies, as the companies may have overlapping target portfolios, but typically do not share assay protocols.

    The data preparation process 
    was executed by the pharma partners on premise, according to a data preparation manual~\cite{melloddy}. The first stage consisted of extracting the data from the individual data warehouses into a standardized file format. Here the partners selected the assays to include, applied unit conversions and scaling, and assigned the correct assay type. The second stage of the data preparation was done using a shared code package called MELLODDY Tuner~\cite{mtuner}, which was built on the open source cheminformatics toolkit RDkit~\cite{greg_landrum_2021_5242603}. MELLODDY Tuner processes the chemical structures and performs structure standardization, calculation of the Morgan Fingerprint representation~\cite{Rogers2010} used as features for machine learning, and assignment of train-test fold split~\cite{Simm2021}. The following four processing steps were performed on the activity data: (1) plausibility checks on activity values, (2) replication of aggregation, (3) identification and application of classification thresholds for classification models and (4) filtering of tasks by data volume quorum to ensure sufficient data for training and robust performance metric calculation. 
    
    Finally, the data was written out in sparse matrix format required for the machine learning algorithm. At this stage, only the data necessary for machine learning was retained, namely the structure feature matrix, the fold allocation information, the matrix with the task labels, and a list of task weights. Different data sets were created this way: a classification data set (CLS) with only classification tasks, a classification data set including auxiliary tasks (CLSAUX), a regression data set with only regression tasks (REG) and a hybrid data set with both classification and regression tasks (HYB). The use of MELLODDY tuner and the jointly-approved Data Preparation Manual ensured the consistency and compatibility of the data prepared by each partner. 
    
    The chemical structures originally exported from data warehouses in the first stage, as well as any assay metadata, such as assay names or targets, were removed. This ensured that only the minimally required data set was present on the machine learning platform. On the platform the tasks were identified by the column index in the label matrix, and only the pharma partners kept on their end the metadata file allowing to map back model predictions to the original assays. In total, the pharma partners included data from $\sim$100 million measurements covering over 40,000 assays and $\sim$20 million compounds for the main prediction tasks. Added to this this were $\sim$2 billion measurements from auxiliary assays.  
    
\subsection{Federated Learning Formulation}\label{sec:federated_form} 
    

    The principle goal of federated learning is to train a global model by minimizing a global objective function $\mathcal{L}_{t}$ which represents the weighted sum of the local objective functions $\mathcal{L}_{p}$ of each partner contributing with private data $\mathrm{D}_{p}$ to the global federated model:
    \begin{equation}
    \begin{aligned}[b]
        \min_{\theta}{\mathcal{L}_{t}(\mathrm{D},\theta)}\;\;\text{with}\;\;\\ \mathcal{L}_{t}(\mathrm{D},\theta)  = \sum_{p=1}^{P}{w_p \mathcal{L}_{p}(\mathrm{D}_{p},\theta_{p})},\;\;
        \bigcap_{p=1}^{p}\theta_{p} \neq \varnothing.
    \end{aligned}
    \end{equation}
The data $\mathrm{D}$, with $\mathcal{X}$ as feature space, $\mathcal{Y}$ as label space and $\mathcal{I}$ as sample ID space, of the $P$ partners can have several distribution characteristics. In the work of~\citet{yang2019federated} a categorisation is proposed for federated learning (FL) depending on the distribution characteristics of the data, e.g. horizontal and vertical FL. In the MELLODDY use case, the feature space $\mathcal{X}$ is the same for all partners however in general the label space $\mathcal{Y}$ and sample ID space $\mathcal{I}$ would differ:
        $\mathcal{X}_i = \mathcal{X}_j,\quad \mathcal{Y}_i \neq \mathcal{Y}_j,\quad \mathcal{I}_i \neq \mathcal{I}_j,\quad \forall \mathrm{D}_i,\mathrm{D}_j, i \neq j$.
In practice we expect some (slight) overlap among the participating partners in label space $\mathcal{Y}$ and sample ID space $\mathcal{I}$ in order to enable federated transfer learning.

    A possible scheme for learning a global model in a federated way would be by using Federated Averaging(FedAvg)~\cite{mcmahan2017communication} proposed by Google. In this scheme, a global model is trained based on iterative averaging orchestrated by a central server where the number of participants is typically large. It concerns a cross-device FL setting where the learning takes place remotely for multiple iterations instead of updating the model every iteration to decrease communication at the cost of model performance.

    In contrast to cross-device FL, MELLODDY is a cross-silo FL scenario, which involved small number of participants (simply 10 pharmaceutical companies) who participated and contributed to the federated model each learning iteration. The scheme used in MELLODDY was based on secure aggregation for federated learning~\cite{bonawitz_practical_2017} where the gradients of all participating partners were aggregated each iteration in a cryptographic, secure way to update the global model (see Figure~\ref{fig:sec_aggr}) instead of sending the model weights and averaging them as in FedAvg. 
    
    \subsection{Models}
    \subsubsection{SparseChem: Base Model for MELLODDY}
    
    Multi-task learning~\cite{caruana1997multitask}, a paradigm also often used in the drug discovery~\cite{dahl2014multi, simoes2018transfer} field for  for quantitative structure–activity relationship (QSAR) models, enables the joint training of a machine learning model where related tasks are involved. 
    SparseChem~\cite{arany2022sparsechem} offers an easy and efficient way to train an industry-scale (millions of input compounds) multi-task QSAR deep neural network models with high-dimensional sparse input features. SparseChem supports classification, (censored) regression and hybrid (both classification and regression) models and were used as base model for MELLODDY for the single partner baseline models but also for the global model from the federated platform as visualised in Figure~\ref{fig:headtrunk}.

\subsubsection{Federated Model: Private Head and Common Trunk}\label{sec:priv_head_common_trunk}
\begin{figure}[!t]
       \centering
      \includegraphics[scale=0.53]{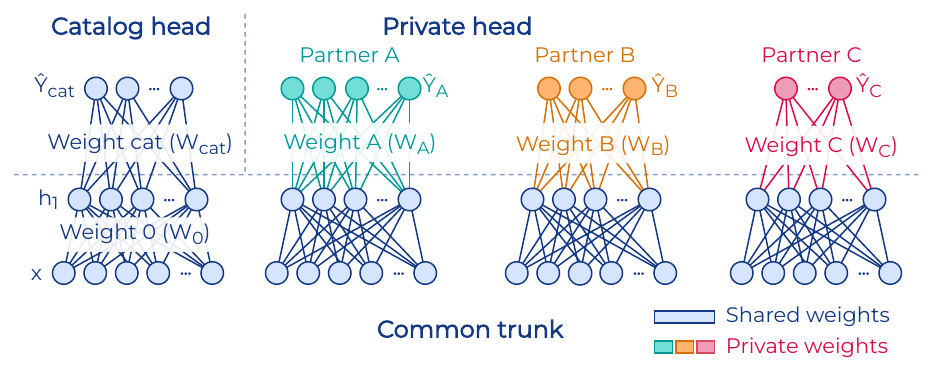}
   \caption{
   The SparseChem models in this figure have one hidden layer: $\mathbf{h_1}=\sigma(\mathbf{W_0}\mathbf{x})$ where $\sigma$ can be a chosen non-linearity (e.g. tanh, relu). The output $\mathbf{\hat{y}_n}=\mathbf{W_n}\mathbf{h_1}$ is a vector where each element represents a different task.}
       \label{fig:headtrunk}
    \end{figure}

The contributing partners to the global federated model can all have different tasks which should be kept private to each partner; this is therefore split into a private head and a common trunk as visualised in Figure~\ref{fig:headtrunk}.
During execution of the federated MELLODDY platform, the weights of the common trunk were trained jointly by all contributing partners through the application of federated learning using secure aggregation of the individual private gradients.
The weights for the private head however remained private for all individual partners; likewise, no communication was needed while training, as the private gradients were used directly to update the private head.
After federated training, the resulting model for each partner consisted of stacking the common trunk (which should be the same for all partners) with the private head (which is different and private for all partners).
\subsubsection{Catalogue Fusion Model}
As previously mentioned, tasks may vary across partners and should remain private. However, the algorithm and platform also provides the opportunity for partners to agree on some tasks to be shared among all or a subset of partners. This holds the potential to enhance the model performance, as for these tasks the model weights would be shared completely as visualised in Figure~\ref{fig:headtrunk} and trained jointly. In the context of MELLODDY and drug discovery for example, the consortium could allow individual partners to contribute to a shared head where the tasks would represent commercial assays such as safety panels~\cite{bowes2012reducing} performed at contract research organisations (CROs). The catalogue (shared) head would only be shared with partners contributing to the catalogue tasks.

\subsection{Risk Analysis} 
\label{sec:risks}    


    The purpose of Risk analysis is to identify and mitigate risk events caused by stealing or manipulating confidential information
    which can have potentially negative impact on the benign participants. To make this analysis as comprehensive as possible, we use a systematic approach detailed in \cite{pejo2022collaborative}. 
    
    \subsubsection{Methodology}

     The initial step of any risk analysis is to define the perimeter, i.e. the actors and the data involved. These are usually referred as Risk Sources and Assets, respectively. The former is a person or non-human entity that can cause a Risk, accidentally or deliberately. The latter are the confidential or private data that the Risk Sources aim to learn or manipulate, thereby causing Risk. The two primary Risk Sources are the participants (pharma companies) and the coordinator (aggregator) server, while the assets are the Chemical Fingerprint, the Targets, the Bioactivity, and the Model. 
    
    A Risk represents the goal of the Risk Source which is to infer confidential information about one of the asset.
    A Threat is a sequence of actions (or attacks) carried out by a Risk Source to realize one or more Risks. A Risk is characterized by its impact (which is measured by its associated negative impact called Severity), while a Threat is characterised with its feasibility (which measures the technical difficulty for a Risk Source to realize the Threat). Combining the feasibility and the success probability of these attacks form the likelihood of the Risk, and this further combined with the Severity allows for identifying the most dangerous risks which should be mitigated. 
    
    For example, a possible risk is that a participating pharmaceutical company $A$ (Risk source) learns that a particular chemical compound (Asset) is used by another participant $B$ from the model updates sent by $B$ for aggregation. A specific threat realizing this risk is that $A$ captures the model update of $B$ and launches inference attack  on the captured update. This risk has maximum severity since $B$'s business strategy may be disclosed as a result of a successful attack. The goal of the analysis is to estimate the success probability and feasibility of such an inference attack.
    
    \subsubsection{Analysis}
    
    The platform architecture implies that the code of the clients and the server is audited and verified. Since malicious manipulation of the training data can result in a larger accuracy drop for a malicious party than for the others \cite{pejo1together}, there is no real incentive for active attacks (e.g., poisoning, back-doors \cite{goldblum2020data}) as long as the adversary also needs good model quality. Consequently, only honest-but-curious adversaries are considered which legitimately participate in the learning protocol. They follow the learning protocol faithfully but also attempt to infer  confidential information about the assets. In this setup, passive attacks are still possible: the trunk is shared among all participants, therefore, it is necessary to understand whether its output (i.e., trunk activation values) or its updates (i.e., gradients) leak any information about the assets. 
    
    \subsubsection{Threats}
    There are many attacks for federated learning; we give a non-comprehensive list below. For more detailed surveys, we refer the reader to \cite{liu2021machine}. 
    
    \begin{itemize}
        \item Model inversion attacks aim to reconstruct a representative training sample of a class \cite{fredrikson_model_2015}, i.e. a record that is similar to all records belonging to a class. 
        \item Membership inference attacks aim at inferring if a certain record was part of the target model's training dataset \cite{hu2021membership}. The most common techniques rely on shadow models \cite{shokri_membership_2016} and can exploit overfitting \cite{pyrgelis_knock_2017}. 
        \item Reconstruction attacks take membership inference attacks another step forward by reconstructing complete training samples \cite{zhu_deep_2019}. 
        \item Property inference attacks aim to infer properties of training data that are independent of the features that characterize the classes of the joint model \cite{ganju_property_2018}. 
        \item Model extraction attacks arise when an adversary obtains black-box access to some target model and attempts to learn a model that closely approximates or even matches the original model \cite{TramerZJRR16}. 
    \end{itemize}

    Despite the wide range of attacks, for our analysis, we focus on membership inference attacks, when the adversary checks if a given compound has been used to train the common model or not. 
    If this elemental attack succeeds, that flags information leakage, while if it does not, that can be a solid empirical argument that other attacks (that potentially leak more information) would probably fail as well. The accuracy of the membership inference is above 90\% when it is launched on the model update of a single participant \cite{pejo2022collaborative}, which means that the likelihood of the corresponding risk is very large supposing the adversary (e.g., the server) can access the update sent by this participant. \medskip
    
    \subsubsection{Mitigations}
    In general, there are legal, organisational, and technical controls.
    Here, we focus only on technical measures.
    A handful of technical mitigation techniques have been proposed against membership inference attacks such as regularization and hyperparameter tuning  \cite{yeom_privacy_2017}. Differential Privacy \cite{pejo2022guide} can also be applied to provide provable privacy guarantees but comes with unacceptable accuracy degradation in our case \cite{pejo2022collaborative}. Besides these, secure aggregation \cite{AcsC11,bonawitz_practical_2017} is a widely used technical control to prevent access to the individual model update while preserving the functionality of the aggregation protocol. In other words, only the pharmaceutical companies learn the aggregated model, and the server performs aggregation without learning anything about the assets. 
  Although a malicious participant can still learn that one of the parties used a specific training sample by launching a membership attack on the aggregated model update or the output of the trunk, it cannot attribute this membership information to any specific party as long as all of them participate in every single round.  
    
    
    Nonetheless, parties may join or leave during training, thus, via a successful membership test on the aggregated model update or the output of the trunk model, the attacker can attribute a training sample to the leaving/joining party. 
    This is a differentiation attack that takes advantage of the change in the coalition. Note that this can be mitigated via legal controls, e.g. by allowing the join of leave of parties only in groups. Altogether, secure aggregation coupled with legal controls mitigate most inference attacks without the degradation of model quality such that the remaining risks become acceptable \cite{pejo2022collaborative}. 
   

    \subsubsection{Contribution Scoring}
    Since secure aggregation prevents the disentaglement of the participants' contributions, it also makes measuring their usefulness within the collaboration more difficult. This additional requirement is often critical when monetary gain is involved, and determining the reward distribution is necessary in case the collaboratively trained model is sold. The silver bullet for contribution score computation is the Shapley value \cite{shapley1953value}, but unfortunately, it is not feasible to compute in many real-world scenarios. Hence, most prior works only approximate that without privacy in mind: they assume access to individual datasets or the corresponding gradients \cite{wang2020principled}. Computing contribution scores privately is a largely unexplored direction, the only works considering this setting are \cite{pejo2020good} and \cite{pejo2021measuring}. The former does not apply to our use case, as it assumes dynamic changes in the training coalition. However, the latter could be adopted with care: the participants compute their scores, thus, verifiable computation techniques should be applied to prevent dishonest reporting. 
    
\section{Platform Blueprint} 
    
    \subsection{Infrastructure}\label{sec:infrastructure}
A cloud setup was selected as the infrastructure for the privacy preserving federated machine learning. To this end, a multi-account setup at Amazon Web Services (AWS) was developed to address the computational demands of MELLODDY while, at the same time, providing a secure environment for the sensitive information of the pharmaceutical partners of the project. The AWS organisation contained all AWS accounts, including two central accounts and a dedicated AWS for each pharmaceutical partner. 
  
 The first central account, the ``orchestration account'', hosted a Kubernetes cluster providing a version-control system, a setup for a public key infrastructure (PKI), and services for cost aggregation and cost reporting. This account was managed by Kubermatic. 
   The second central account, the ``central ML account,'' was utilised to provide the core components of the federated machine learning framework, namely the model dispatcher. The model dispatcher was deployed on a Kubernetes cluster. This AWS account was managed by Owkin.

   Each pharmaceutical partner owned a dedicated AWS account containing multiple services to store the sensitive data, to pre-process the data, manage the compute resources in the account and finally to perform the federated machine learning. The key services in the pharmaceutical partner accounts can be described as follows: The sensitive data was stored in a secured S3 bucket and prior to the upload, internet access is removed from both pharma private VPC and shared VPC. A Kubernetes cluster was deployed using EC2 instances, i.e. the control plane was hosted by three instances managing an adaptively adjustable number of worker node. The worker nodes of the Kubernetes cluster were located in a ``shared subnet'' within a peered VPC (virtual private cloud) to enable high-bandwidth network connectivity between local resources and the model dispatcher.  A ``console'' server was utilised to deploy and manage the Kubernetes cluster and additionally, to host a CLI (command line interface) which was---in turn---used by the partners to adjust the number of worker nodes, to trigger the pre-processing of data, and to initiate the federated machine learning runs. The AWS services were monitored by CloudWatch not only for health status but also for security issues. The infrastructure was deployed using the tool Terraform~\cite{terraform}.

    The entire MELLODDY platform, including the infrastructure setup at AWS, underwent a thorough security audit by an external auditor (i.e. Cirosec); findings and code were revised by each pharma company's security experts and the platform was ultimately signed off by all partners. The successful audit is testament that the setup is capable of handling sensitive classified data in the context of drug discovery. To allow changes of the setup after the audit, a review process had been implemented. That is, every change of the underlying code had to be reviewed by at least three pharmaceutical partners for security issues before being deployed to production.
    
\subsection{Application Layer} 

    \subsubsection{Platform}

\begin{figure}
        \centering
        \includegraphics[scale=0.0965]{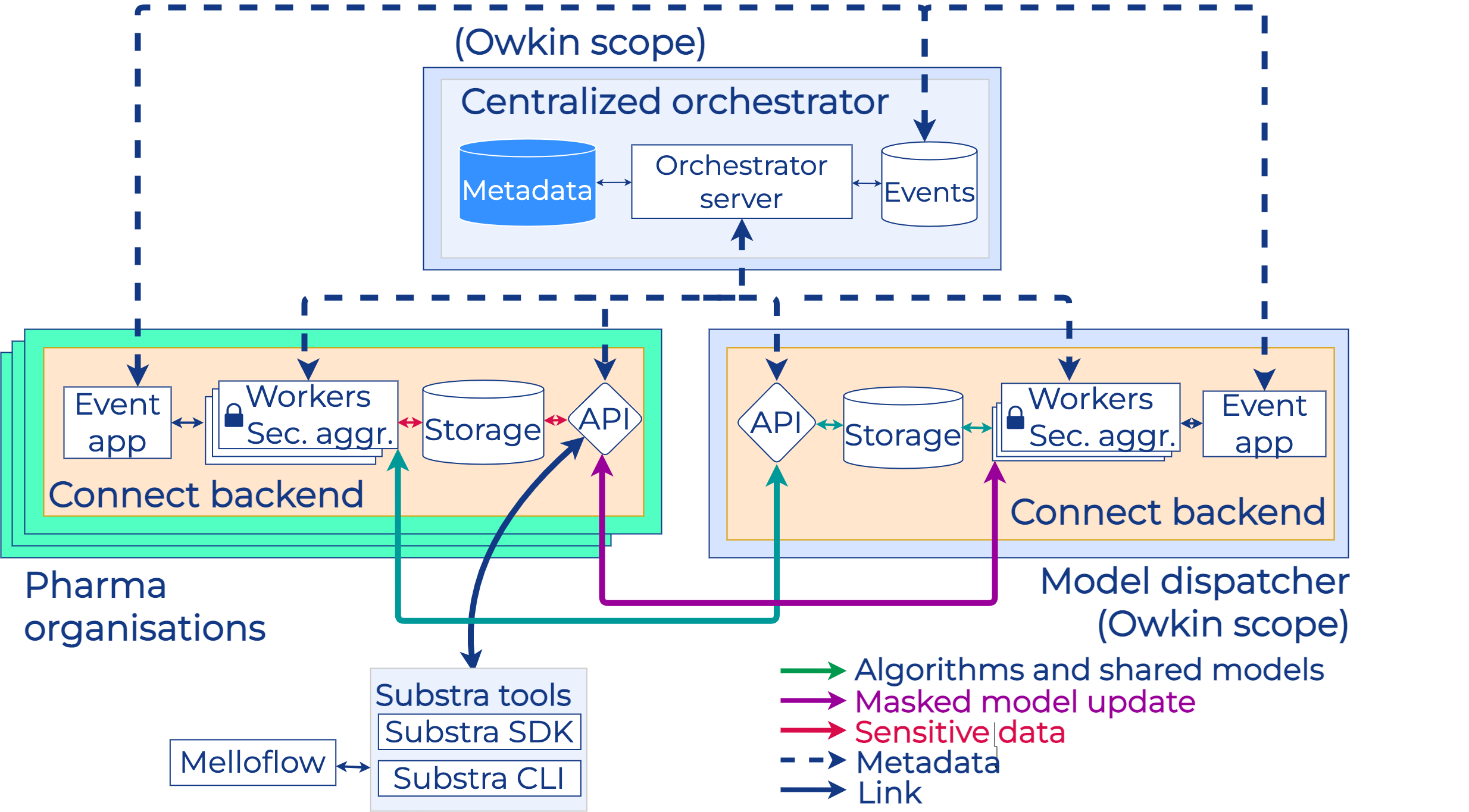}
        \caption{Overview of the deployed platform}
        \label{fig:platform_overview}
    \end{figure}

    The platform consists of a set of interconnected organizations. Figure \ref{fig:platform_overview} represents the main components of a deployed platform in two organisations: a pharma partner and a central aggregation one.

    The application layer of MELLODDY relied heavily on Owkin Connect (based on Substra~\cite{substra}, open-source software mainly developed by Owkin), an application to train and evaluate machine learning models on distributed data sets without centralising the data or compromising the privacy of the data  maintaining traceability of training and evaluation operations. It supports simple training and evaluation scheme, such as training a model on data in a center A and evaluating the model on data in a center B, or more complex federated learning (FL) schemes, such as Federated Averaging. It also enables both horizontal FL and vertical FL, as well as multi-partners multi-task learning.

    The platform uses the following elementary components:

    \begin{description}
        \item[Federated Learning (FL) Orchestrator] is responsible for the orchestration of ML tasks on distributed datasets: it distributes the tasks to the workers of the different organizations. It stores non-sensitive metadata of assets of Connect, makes it possible to verify the integrity of assets and ensures that permissions on assets are respected. The orchestrator can be a centralized component or used in a decentralised mode through a Distributed Ledger Technology (DLT)~\cite{rauchs2018distributed}.

        \item[Connect-Backend] is a core component in each organisation: its REST API is the main entry point to interact with the platform. It also handles the storage of assets like algorithms, data samples and models. One of its main subsystem is the compute engine, where algorithms and data samples meet to create added value. The compute engine can scale horizontally to leverage multiple compute resources (CPU/GPU/memory).

        \item[Connect Interfaces and Libraries] used to interact with the connect backend either by data scientist or IT operational contacts. The most notable ones are the frontend, allowing to monitor assets and compute plan execution; there are also several Substra libraries (substra-tools, substra SDK \& CLI) to simplify algorithm definition and API interaction.
        
        \item[Melloflow]\label{comp:melloflow} is a library and CLI providing mini-batch generation and data set, algorithm and hard coded compute plan registration in the context of the MELLODDY project. One of the central library is the secure-aggregation used for federated learning of models.

        \item[Deployment Artifacts and Tools] regrouping both Kubernetes manifests to deploy Owkin Connect, melloddy CLI and its associated server providing on-demand manifest generation for pharma operators.
    \end{description}

    \subsubsection{Compute Plans}

    The ML experiments executed on the platform were registered and executed as compute plans, a concept specific to Connect. A compute plan is a directed acyclic graph of tasks. A task used for training is defined by an algorithm, the data it is executed on and the inputs from its parent tasks. A task used for evaluation is defined by a set of metrics, the train task it evaluates and the data used for the evaluation. An aggregation task is defined by an algorithm and the inputs from its parent tasks.

    The algorithm contains the code executed by the task and the description of the environment of the execution, via a Dockerfile. The metrics define how to get a score from predictions and the ground truth.

    In the context of the MELLODDY project, the data was registered under the format of samples, each sample corresponding to one mini-batch of an epoch. The mini-batches were generated from the output of the MELLODDY Tuner~\cite{mtuner} library using the Melloflow library.

    On the platform, the data does not leave the organisation it was registered on. Each task can only be executed on the data from one organisation and the execution takes place on that organisation. The other assets (algorithm, metric and task inputs) move from one organisation to the other as needed.

    The algorithms and metrics are shared publicly to all organisations. The task input and outputs may contain sensitive information and as such are kept private. The training tasks have two outputs: (1) the full model which does not leave the organisation and (2) the model metadata (e.g. trunk gradients), that is shared with the central aggregation organisation. This model metadata is encrypted using a secure aggregation scheme.

    The aggregation task is executed on the central organisation, taking as input the model metadata from other organizations and returning an aggregation of the metadata, which is sent to each organization. The test task outputs a list of metrics, which are anonymised and showed publicly on the frontend.
    
    \subsubsection{(Sparse) Secure Aggregation}
    
    The Secure Aggregation protocol is utilized to prevent attribution of any inferred information from the shared trunk model. Within Connect it is based on~\cite{AcsC11,bonawitz_practical_2017} with two changes: the secret sharing is disabled, but a common secret mask shared among all participants is utilized. The former is not required, since all partners must stay connected all the time in order to prevent differential attacks. The latter is essential to prevent the aggregator from accessing the result of the aggregation: in addition to the pairwise masks, each participant adds another secret mask to their model update that is unknown to the server and can be removed only by the participants. Hence, the aggregator can perform aggregation without learning anything about the model updates \emph{and} their aggregate. 

        
    
        
    
    Although the computationally heavy secret sharing is removed, the new protocol still incurs a significant communication and computational cost, as random keys must be generated and added to each gradient value of all participants in each round. Yet, a considerable portion of the gradient update values is zero because the training data within a batch correspond to a small number of tasks and the input chemical fingerprints are also sparse. Therefore, the gradients are sparse as well, and compressing the model update before encryption can increase efficiency with presumable minor accuracy drop. Moreover, these techniques also mitigate confidentiality risks to some extent \cite{pejo2022collaborative}. 
    
    Unfortunately, secure aggregation hides the  location of non-zero coordinates, and hence participants do not know which gradient values are non-zero at the other parties.  
    In MELLODDY, a simple and cheap approach is followed, that is, every participant sends the gradient values at exactly the same random subset of coordinates for aggregation. In particular, 
    each participant first selects the same random subset of coordinates uniformly at random using a common secret seed, then  it sends the encrypted gradient values only at these coordinates for aggregation. After decrypting the aggregate, only the aggregated coordinates are updated, the rest remains unchanged.
    
    
\subsubsection{Partner Weighting}

The platform supports runs with different schemes of relative weighting of partner contributions within each training iteration of the federated run. Partner weighting is enabled by scaling each gradient of each partner before aggregating. For default partner weighting the gradients of all partners are scaled by a constant so that each data point is equally weighted and partners with more data contribute more to the aggregated gradient. Other possible schemes are to scale the gradient depending on the mini-batch size or number of non zeros in the mini-batch per partner. This would allow partners to contribute more equally to the aggregated gradients. In the work of~\citet{melloddy} it was however found that default weighting was more beneficial for the federated model.

    \subsection{Platform Tools} 
   


    \subsubsection{MELLODDY Federated Learning (FL) Simulator}
    
 During the MELLODDY project, many options were considered to improve the performance of the model~\cite{melloddy}. In order to quickly assess each option, the partners conducted single partner studies. In these studies, they used the Melloflow~\ref{comp:melloflow} library to simulate FL experiments locally on public data and their own data, split into n virtual FL partners. This allowed them to get preliminary results for each option, and decide whether to integrate the option into the final experiment.
    Melloflow, used on top of Owkin Connect' local backend,  can simulate the FL experiment by running the tasks locally with Python subprocesses or Docker containers on the same development environment.
    
    \subsubsection{MELLODDY Predictor}
    
    MELLODDY Predictor~\cite{mdypredictor} is an open-source Python package made for external data scientists without high knowledge of the MELLODDY stack to perform predictions on new data easily from the models produced during the yearly runs. It is built on top of MELLODDY-Tuner and SparseChem to manage both data pre-processing and model inference steps. It is flexible enough to handle multiple models and data size, and predict on subset on tasks. 
    
    \subsubsection{Model Fusion}
    Model fusion is a processing option designed to increase the overall performance. It enables selection of the best model per task instead of a single best average performer. To this end, each partner can select the group of already trained models from the pool of single- and multipartner models. The best model for each task is then selected based on the separate dataset and the performance is measured based on the held-out test set~\cite{melloddy}.

\subsection{Operational Model}
    
    The MELLODDY model required the collaboration of partners with different roles in the consortium. These different roles were reflected in the platform as various tasks to perform and security permissions.
    
    \subsubsection{Decentralised Administration}
    
    
    A decentralised approach had to be implemented to meet with the strong confidentiality requirements of MELLODDY and to allow for the industry partners' private and sensitive data sets to be kept in their respective private IT environments. This also required specific processes and operation sequences to be developed. In particular: (1) partners set up and maintained their own IT platform component environments, (2) each partner was represented for IT operations by an "Operational Contact" in the project, and (3) a detailed coordination approach was elaborated.


    This decentralized approach raised a number of challenges and difficulties: (1) as only the operational contact of each partner was able to access the IT environment of a given industry partner, remote assistance without access was set up in order to resolve bugs, (2) a fine-grained planning of operations was necessary to take into account working hours and time zones, and (3) an error by one partner can potentially result in numerous operations for all other partners.
    

\subsubsection{Different Phases of Operation}
Operations were split into 3 phases: (1) Phase 1 was used for hyper-parameter tuning of machine learning models. In this phase only 60\% of the data was used to train the machine learning models, while 20\% of the data was used to evaluate the models and the other 20\% is left out. (2) In phase 2 the best hyper-parameters were selected from phase 1 and the machine learning models were retrained using the 80\% of the data and evaluated on the 20\% left out data in phase 1. (3) In phase 3 100\% of data was used to train machine learning models using best hyper-parameters from phase 1. The performance of the models could not be evaluated in phase 3 (100\% of data is used as training data) but we assume they perform better compared to phase 2 models as more data was used to train them.

\subsubsection{Application Use}
The MELLODDY project spanned 3 years, with a run taking place each year using the most recent version of the platform. The performance increased (number of compute plans) each year and in year 3 the platform supported a run of 219 compute plans on the 4 different data sets (see Data Section~\ref{sec:data}) over the 3 different operational phases. More precisely, 34 compute plans ran on CLS data set, 47 on CLSAUX data set, 62 on REG data set and 76 on HYB data set. A total of 1.189M US\$ compute budget was spent in year 3 with a maximum of 32 workers in parallel per partner. 
These 32 workers were distributed over 2 parallel platforms. A single epoch of a compute plan, depending on model complexity and number of data points, would take 1.7 hours to 8.1 hours using 50 mini batches or 2.8 hours to 12.9 hours using 80 mini batches. A compute plan would typically run around 20 epochs to reach model convergence.
During application use the main bottlenecks encountered were communication overhead and memory usage. The communication overhead is inherent to federated learning and increases with number of mini batches but also number of model weights. The memory usage increases when model size increases (e.g. number of tasks) or mini-batch size (e.g. data-points). With these bottlenecks in mind our choice for AWS instances resulted in \textit{c5n.18xlarge} (192GiB Memory, 100Gbps Network bandwidth) for the central aggregating node and \textit{r5n.2xlarge} (64GiB Memory, up to 25Gbps Network bandwidth) for the workers of the contributing partners. 

\begin{figure}
        \centering
        \includegraphics[scale=0.3]{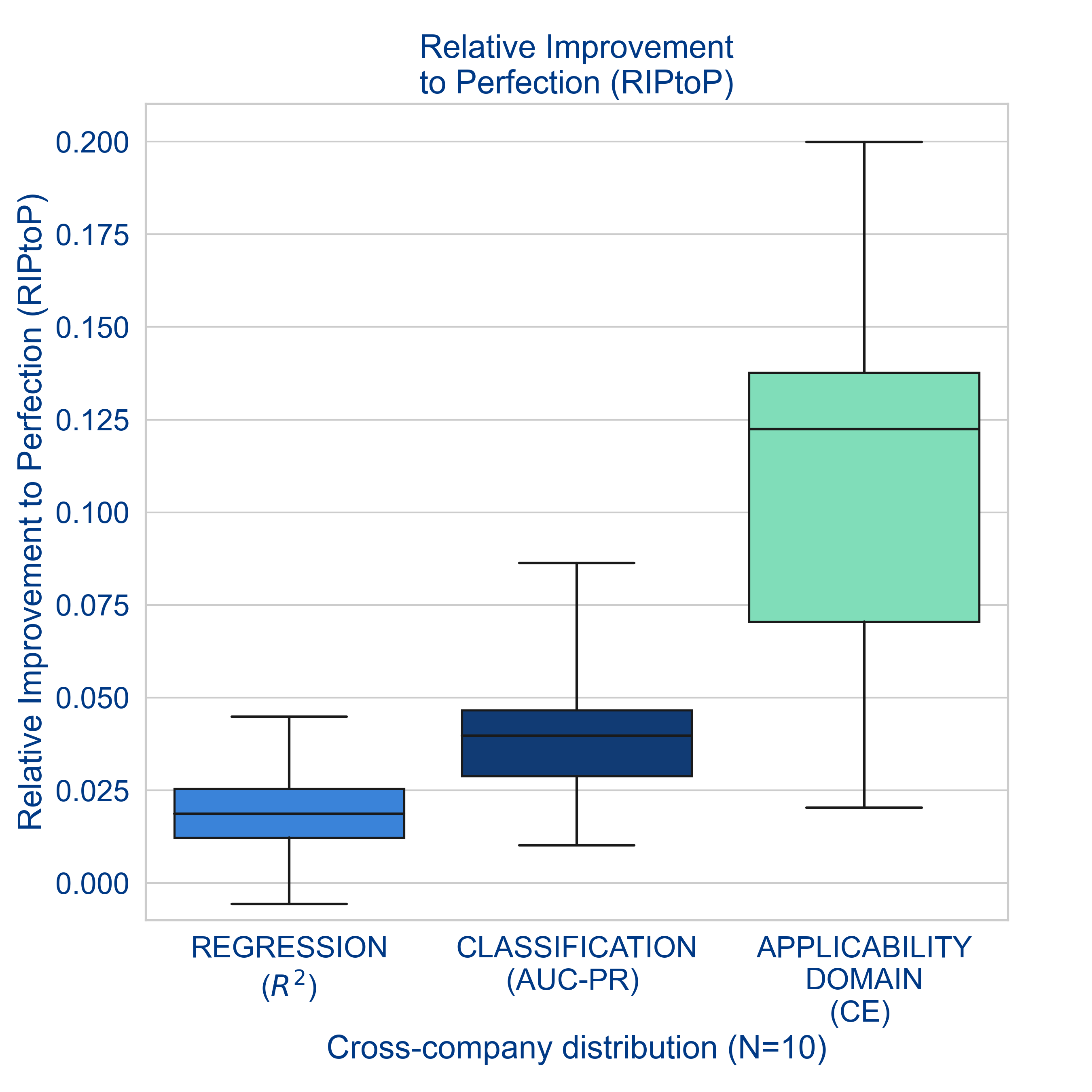}
        \caption{The multi-partner models trained on the platform resulted in a relative improvement for all partners compared to single partner model for all metrics: (1) the AUC-PR for classification, (2) the $R^2$ for regression, and (3) the extension of the domain of applicability as the delta median conformal efficiency for classification.}
        \label{fig:y3_results}
\end{figure}
\section{Results}

The deployed application at scale resulted in improvements across all pharmaceutical partners in the predictive performance of collaboratively trained models over single partner models as shown in Figure~\ref{fig:y3_results}. For the classification models produced by the platform the primary evaluation metric was AUC-PR on average across 100.000+ ML learning tasks representing 40.000+ concentration-response assays. For the regression models the metric reported was the $R^2$. Lastly, the extension domain applicability (AD) measured as the delta median conformal efficiency for classification~\cite{conformal} means that the model can support navigation of a broader chemical space previously unknown to that partner. On Figure~\ref{fig:y3_results} the relative improvements to perfection (where single partner model performance is 0 and perfect model is 1) for the three metrics (AUC-PR, $R^2$ , CE) are visualized for the cross-company distribution ($N=10$) aggregated over all tasks. Across all pharmaceutical partners, federated models were typically 4\% better at categorizing molecules as either pharmacologically or toxicologically active or not active. The typical multi-partner model also showed a 10\% increase in its applicability domain, its ability to yield confident predictions when applied to new types of molecules. Finally, the federated models were typically 2\% better at estimating values of toxicological and pharmacological activities. Performance improvements were more prominent for the subset of assays relating to pharmacokinetics and toxicology and for assays with ongoing data acquisition. For further details, we refer to the work of \citet{melloddy}. 

Collectively, these results show improvements to predictive models that support the drug discovery process holding the potential benefits for the discovery of new drugs. Models that more accurately predict molecules' pharmacological and toxicological activities better support the decision-making process of which candidate drug molecules to make and test. All ten pharma partners attest to observing benefits for living and ADME assays and aim to utilise the models in their internal pipelines. 
\section{Conclusion and Next Steps}
The ready to use platform described in this work demonstrates that federated learning for drug discovery is possible on industry scale. It enabled a groundbreaking collaboration without sharing data in a industry where data confidentiality is high priority. The platform is deployed easily on a cloud infrastructure using a AWS multi-account setup and has already run for 3 years in production.  We also explained how a decentralised administration works as a organizational model in a real case federated setup. Many options remain to be explored in future work such as sparse secure aggregation or partner weighting as well as post processing tools of the platform like model fusion. There is opportunity for the platform to be further optimised so that current bottlenecks like communication and memory limitations are reduced. There is interest to use the platform beyond MELLODDY scope and formalities and initiatives are ongoing. 
Finally, code related to the MELLODDY project is made available on GitHub~\cite{mellodycode}.

\section{Acknowledgments}
This project has received funding from the Innovative Medicines Initiative 2 Joint Undertaking under grant agreement N° 831472. This Joint Undertaking receives support from the European Union’s Horizon 2020 research and innovation program and EFPIA. YM, AA, JS, AF and MO are affiliated to Leuven.AI and received funding from the Flemish Government (AI Research Program). 

{\small \bibliography{aaai22}}

\end{document}